# Discovering long term dependencies in noisy time series data using deep learning

Kurochkin Alexey
kurochk1n.alexey@yandex.ru

**Abstract**

Time series modelling is essential for solving tasks such as predictive maintenance, quality control and optimisation. Deep learning is widely used for solving such problems. When managing complex manufacturing process with neural networks, engineers need to know why machine learning model made specific decision and what are possible outcomes of following model's recommendation. In this paper we develop framework for capturing and explaining temporal dependencies in time series data using deep neural networks and test it on various synthetic and real-world datasets.

## 1. Introduction

Machine learning is revolutionising many industries such as chemical, metal and mining, pulp and paper, oil and gas. Ability to discover and forecast complex nonlinear dependencies in sensorial time series data leads to revealing hidden value for industrial companies. In general, where are three main ML use cases in heavy industry: predictive maintenance, quality control and production optimisation. For example, such tasks as anomaly detection during steel pipe welding, prediction of electric centrifugal pump breakdown or prediction of paper sheets breaks in paper production are related to predictive maintenance. Quality control tasks are prediction of P2O5 concentration in fertilising production, forecasting of coal humidity and ash factor. Optimisation of petroleum distillation using machine learning for controlling oil refinery leads to more stable product quality with less energy consumption.

Deep learning is part of machine learning which is famous for its exceptionally good performance in computer vision natural language processing, reinforcement learning and time series modelling. Deep learning is also successfully used for solving the above industrial tasks. However, deep neural networks behave as black boxes during operation and their predictions are not explainable due to high number of model hyper parameters and overall complexity. At the same time, engineers who control complex manufacturing process need to know why model made a specific decision, what actions they should take and estimate risks and outcomes when following recommendations of the model. For example, if anomaly detection model predict failure, decision about production line shutdown must be made based on model confidence in its decision and explanation of what equipment of production line behaves abnormally and when this anomaly behaviour started. For these reasons, for successful integration of deep learning in manufacturing process model must not only perform well in terms of standard prediction metrics, such as mean squared error or accuracy, but also be explainable.

In addition, modern production facilities are overwhelmingly complex, involving hundreds of different equipment units and thousands of sensors. In practice it is often impossible for human to account contribution of every production segment at every timestamp into final product quality. Sometimes process engineers do not know all details of manufacturing process and perform grid search on production parameters with the aim of finding parameter set which best matches the de-sired results. Such control may be non optimal, and knowledge of all equipment impact on final product quality at given timestamp can significantly boost the manufacturing performance.

These problems emphasise the essential importance of creating temporal knowledge graph from raw multivariate time series data. Also, in many process industries several hours are passed between loading raw materials and estimating quality of final product. Production performance vitally relies not only on short term but also on long term dependencies between production facilities. For instance, in case of gold extraction from raw ore, data collected from adsorption equipment, which is one of last components of production line and affect extraction percentage last few hours before extraction measurement, is of same importance as data collected from ore mill and centrifugal separation equipment which are first components of extraction process and affect extraction percentage long before measurement is conducted.

In this paper we develop deep learning framework which is capable to perform explainable prediction of target variable from raw noisy multivariate time series data and build temporal knowledge graph which accounts for long and short temporal dependencies. Our contribution is as follows:
1) We develop deep neural forecasting models based on state-of-the-art recurrent and convolutional neural networks and asses regression quality on synthetic and real-world datasets
2) We estimate quality of temporal knowledge graphs construction for noisy synthetic time series data with long and short term linear and nonlinear temporal dependencies in terms of precision and recall.
3) We compare performance of recurrent and convolutional models for temporal knowledge graph construction.

Code of our experiments is freely available at https://github.com/KurochkinAlexey/discovering-long-term-dependencies-in-noisy-time-series-data

## 2. Related work

The importance of causal discovery in time series data is emphasised by many authors. In [1] causal inference on synthetic and climate datasets is performed using PCMCI algorithm. In [2] authors used method based on minimising a mutual information-regularised risk with learnable input noise of a prediction model for exploratory discovery of nonlinear directional relations from observational time series. Their experiments were conducted on synthetic, video game and EEG datasets and results outperform other methods in terms of AUC-PR. In [3] authors study causal discovery and forecasting for non stationary time series using a particular type of state-space model to represent the processes. In [4] authors used economy statistical recurrent units for inferring nonlinear Granger causality. Comparing to these methods, our contribution is study of discovering long term dependencies and fast, scalable framework for building causal graph.

Attempts were made towards explainability of deep learning models. In [5] authors used attention-based RNN for interpretable prediction of blood pressure response to fluid bolus therapy. In [6] authors used learned binary masks for interpreting recurrent neural networks prediction for ICU mortality. In [7] authors used multitask Gaussian process and attention-based convolutional network to early predict the occurrence of sepsis in an interpretable manner. In [8] and [9] authors address the problem of deep learning model interpretation with attention weights. In [10] authors use interpretable dual-attention encoder-decoder scheme for time series forecasting. In [11] authors explore

LSTM[13] with variable-wise hidden states and temporal and variable attention to perform interpretable forecasting over multivariate time series data.

Capturing long-term dependencies in sequential data is goal for many recurrent neural network architectures. In [12] authors introduce a special form of recurrent networks which are able to capture long-term dependencies thanks to the stability property of its underlying differential equation. In [14] authors use a novel connectivity structure based on the Schur decomposition to circumvent the exploding and vanishing gradient problem in RNNs and learn long time dependencies. In [15] authors compare generic recurrent neural networks and temporal convolutional networks for sequence modelling.

## 3. Problem formulation

We denote time series as $x^{(i)} = \{x_t^{(i)}\}, i = 1,...N, t = 1,...,T$, where each $x_t^{(i)} \in \mathbb{R}$. Consider following relations between time series:

$$x_t^{(i)} = f^{(i)}(\alpha_{i1} x_{t-\tau_{i1}}^{(1)}, \ldots, \alpha_{iN} x_{t-\tau_{iN}}^{(N)}) + \beta^{(i)} \epsilon$$

where $f^{(i)}$ is some function, $\alpha_{ij}$ is coefficient which regulate strength of causal relation between $i$ and $j$ time series, $\tau_{ij}$ is time lag which controls temporal dependency between $i$ and $j$ time series, $\epsilon \sim N(0,1)$ is white noise and $\beta^{(i)}$ controls signal-to-noise ratio. In all our experiments $\tau_{ij} \in [1,300)$. For given target time series $k$ our goal is to obtain $\hat{\tau}_{ij}$ - estimation of temporal dependency lag, and $\hat{\alpha}_{kj}$ - estimation of causal strength in following form:

$$\hat{\alpha}_{kj} = \begin{cases} 1, & \text{if } \alpha_{kj} \neq 0 \\ 0, & \text{if } \alpha_{kj} = 0 \end{cases}$$

In this paper we do not focus on estimating actual values of $\alpha_{kj}$. After finding factors which affect target time series, for every $j \in \{j \mid \hat{\alpha}_{kj} \neq 0\}$, the goal is to estimate $\hat{\alpha}_{jl}$ and $\hat{\tau}_{jl}, l = 1,...,N$ - values which define dependencies in driving time-series. We limit further discovering of dependencies for each non-zero $\hat{\alpha}_{jl}$ due to high computational complexity of experiments.

By defining set of variables described above temporal knowledge graph of underlying dependencies between time series can be approximately reconstructed.

## 4. Proposed methodology

To discover temporal dependencies affecting target time series $k$, following regression model parametrised with deep neural network is built:

$$x_t^{(k)} = F^{(k)}(X_{t-\tau}, \theta^{(k)})$$

where $X_{t-\tau} = (x_{t-\tau}^{(i)}, \ldots, x_{t-1}^{(i)}), i = 1,\ldots,N$ is multivariate time series and $\theta^{(k)}$ is neural network parameters. After training model, $\hat{\alpha}_{kj}, \hat{\tau}_{kj}, j = 1,\ldots,N$ are obtained through input importance estimation procedure performed on test set:

$$\hat{\alpha}_{kj}, \hat{\tau}_{kj} = ImportanceEstimation(X_{t-\tau}, F^{(k)}(X_{t-\tau}, \theta^{(k)})), j = 1,\ldots,N$$

Then, procedure is repeated:

$$x_t^{(j)} = F^{(j)}(X_{t-\tau}, \theta^{(j)})$$
$$\hat{\alpha}_{jl}, \hat{\tau}_{jl} = ImportanceEstimation(X_{t-\tau}, F^{(j)}(X_{t-\tau}, \theta^{(j)})), l = 1,\ldots,N$$

for every $j \in \{j \mid \hat{\alpha}_{kj} \neq 0\}$.

Again we notice that we limit further discovering of dependencies for each non-zero $\hat{\alpha}_{jl}$ due to high computational complexity of experiments.

The key step to accurate estimation of desired parameters is to build regression model which is capable of detecting complex long term dependencies in time series data. To asses this problem, following neural networks architectures are considered:
1) LSTM[13]
2) GRU[16]
3) Interpretable Multi-Variable LSTM[11]
4) AntisymmetricRNN[12]
5) Recurrent Highway Network[17]
6) Temporal Convolutional Network[15]

## 4.1 Overview of considered network architectures

### 4.1.1 Interpretable Multi-Variable LSTM

In [11] idea of encoding multivariate time series into variable-wise hidden state of recurrent neural network is explored. Such architecture aimed to capture different dynamics in multivariate time series. In addition, mixture attention mechanism distinguish contribution of different variables to prediction and can be used to forecasting interpretation. Default LSTM cell structure is modified as follows:

$$\tilde{i}_t = \sigma(W_i \otimes \tilde{h}_{t-1} + U_i \otimes x_t + b_i)$$
$$\tilde{j}_t = tanh(W_j \otimes \tilde{h}_{t-1} + U_j \otimes x_t + b_j)$$
$$\tilde{f}_t = \sigma(W_f \otimes \tilde{h}_{t-1} + U_f \otimes x_t + b_f)$$
$$\tilde{o}_t = \sigma(W_o \otimes \tilde{h}_{t-1} + U_o \otimes x_t + b_o)$$
$$\tilde{c}_t = \tilde{c}_{t-1} \odot \tilde{f}_t + \tilde{i}_t \odot \tilde{j}_t$$
$$\tilde{h}_t = \tilde{o}_t \odot tanh(\tilde{c}_t)$$

where

$$W_\alpha = [W_\alpha^{(1)}, \ldots, W_\alpha^{(N)}] \in \mathbb{R}^{N \times d \times d}, W_\alpha^{(k)} \in \mathbb{R}^{d \times d}$$
$$U_\alpha = [U_\alpha^{(1)}, \ldots, U_\alpha^{(N)}] \in \mathbb{R}^{N \times d \times 1}, U_\alpha^{(k)} \in \mathbb{R}^{d \times 1}, \alpha \in \{i, j, f, o\}$$

are hidden-to-hidden and input-to-hidden transition tensors,

$$\tilde{h}_t = [h_t^{(1)}, \ldots, h_t^{(N)}] \in \mathbb{R}^{N \times d}, h_t^{(k)} \in \mathbb{R}^d$$

$$\tilde{c}_t = [c_t^{(1)}, \ldots, c_t^{(N)}] \in \mathbb{R}^{N \times d}, c_t^{(k)} \in \mathbb{R}^d$$

are hidden state matrices at timestamp $t$, $N$ represents number of variables in multivariate time series and $\otimes$ denotes tensor product, such that

$$W_\alpha \otimes \tilde{h}_{t-1} = [W_\alpha^{(1)} h_{t-1}^{(1)}, \ldots, W_\alpha^{(N)} h_{t-1}^{(N)}], W_\alpha^{(k)} h_{t-1}^{(k)} \in \mathbb{R}^d.$$

Extending hidden state vectors to matrices leads to improved forecasting quality on multivariate time series data. To boost performance and provide model with interpretation ability, mixture attention mechanism is used, details can be found in [11].

### 4.1.2 AntisymmetricRNN

In [12] problem of learning long-term dependencies in time series data is tackled through connection between recurrent neural networks and ordinary differential equations. Following recurrent cell structure is proposed:

$$z_t = \sigma((W_h - W_h^T - \gamma I)h_{t-1} + V_z x_t + b_z)$$
$$h_t = h_{t-1} + \epsilon z_t \odot tanh((W_h - W_h^T - \gamma I)h_{t-1} + V_h x_t + b_h)$$

where

$$W_h \in \mathbb{R}^{d \times d}, V_h \in \mathbb{R}^{N \times d}, V_z \in \mathbb{R}^{N \times d}, b_z \in \mathbb{R}^d, b_h \in \mathbb{R}^d$$

are network parameters, $I \in \mathbb{R}^{d \times d}$ is identity matrix, $\gamma$ controls strength of diffusion, $\epsilon$ is a step size, $\sigma$ is sigmoid function and $\odot$ is Hadamart product.

Such architecture simulates forward Euler method of solving ordinary differential equation with theoretically proven stability.

### 4.1.2 Recurrent Highway Network

In [17] deep recurrent transitions proposed to learn complex nonlinear transition functions from one timestamp to the next. This is done by using Highway layer in recurrent transformation:

$$s_t^{(0)} = y_{t-1}$$
$$s_t^{(l)} = h_t^{(l)} \odot t_t^{(l)} + s_t^{(l-1)} \odot c_t^{(l)}$$
$$h_t^{(l)} = tanh(W_H x_t I_{\{l=1\}} + R_H^{(l)} s_t^{(l-1)} + b_H^{(l)})$$
$$t_t^{(l)} = \sigma(W_T x_t I_{\{l=1\}} + R_T^{(l)} s_t^{(l-1)} + b_T^{(l)})$$
$$c_t^{(l)} = \sigma(W_C x_t I_{\{l=1\}} + R_C^{(l)} s_t^{(l-1)} + b_C^{(l)})$$
$$y_t = s_t^{(L)}$$

where $y_t \in \mathbb{R}^d$ is hidden recurrent state, $L$ is recurrence depth, $l = 1,\ldots,L$, $I_{\{\}}$ is indicator function, $W_H \in \mathbb{R}^{N \times d}, W_T \in \mathbb{R}^{N \times d}, W_C \in \mathbb{R}^{N \times d}, R_H^{(l)} \in \mathbb{R}^{d \times d}, R_T^{(l)} \in \mathbb{R}^{d \times d}, R_C^{(l)} \in \mathbb{R}^{d \times d}, b_H^{(l)} \in \mathbb{R}^d, b_T^{(l)} \in \mathbb{R}^d, b_C^{(l)} \in \mathbb{R}^d$ are network parameters, $\sigma$ is sigmoid function and $\odot$ is Hadamart product.

Such gates structure provide a more versatile setup for dynamically remembering, forgetting and transforming information compared to default variants of LSTM and GRU, which can be theoretically proven using Gersgoren circle theorem.

### 4.1.4 Temporal convolutional network

Temporal 1D causal convolutions are used as alternative for sequence modelling. There are several advantages of TCN over RNN, such as parallelism, flexible receptive field size and stable gradients[15]. Typical TCN architecture is illustrated in Figure 1.

For 1D sequence $x \in \mathbb{R}^N$ and filter $f: \{0,...,k-1\} \to \mathbb{R}$, the dilated convolution operation $F$ on element $s$ of a sequence is defined as

$$F(s) = \sum_{i=0}^{k-1} f(i) x_{s-di}$$

where $d$ is a dilation factor, $k$ is a filter size

Experiments done in [15] showed that TCN outperforms generic RNN architectures on series of long sequence modelling tasks. To investigate advantages of TCN, we conducted experiments with several architecture design:
1) Output layer attention
2) Layerwise attention
3) Stacking of TCN networks with different input window sizes
4) Bidirectional TCN

For realisation details see https://github.com/KurochkinAlexey/discovering-long-term-dependencies-in-noisy-time-series-data

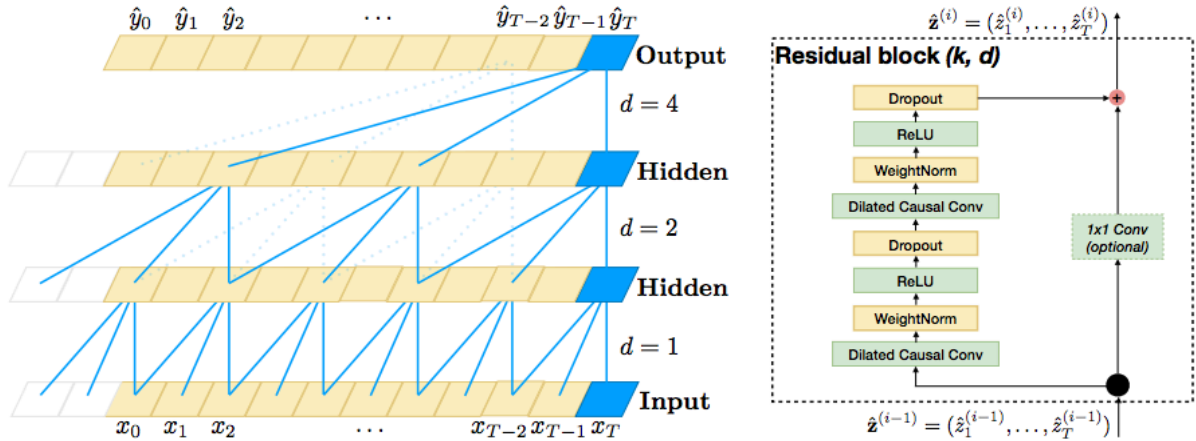

Figure 1. Typical temporal convolution network architecture(taken form [15])

## 4.2 Importance estimation

Once regressor $F^{(i)}(X_{t-\tau}, \theta^{(i)})$ is trained, input $X_{t-\tau}$ importance is estimated through Learned Binary Mask(LBM) approach described in [6]. LBM procedure learns to find binary mask $M_{t-\tau} \in \{0,1\}^{\tau \times N}$, such that after feeding $X_{t-\tau} \odot M_{t-\tau}$ into trained regressor, output prediction is approximately same as prediction with unmasked input:

$$F^{(i)}(X_{t-\tau} \odot M_{t-\tau}, \theta^{(i)}) \approx F^{(i)}(X_{t-\tau}, \theta^{(i)})$$

Binary masks are learned via 2-step process. Firstly, $\tilde{M} = \sigma(x), x \sim N(0, I)$ is learned through minimising difference between masked and unmasked prediction:

$$|y_t^{(i)} - F^{(i)}(X_{t-\tau} \odot \tilde{M}_{t-\tau}, \theta^{(i)})| + \lambda_1 \sum_{t=1}^{\tau} \sum_{j=1}^{N} |M_t^{(j)}| + \lambda_2 L(\tilde{M}_{t-\tau}^{bin}, \tilde{M}_{t-\tau}) \to min$$

where $\lambda_1$ and $\lambda_2$ control mask sparsity, $L$ is cross-entropy loss and:

$$\tilde{M}_{t-\tau}^{bin} = \begin{cases} 1, & \text{if } \tilde{M}_{t-\tau} > 0.5 \\ 0, & \text{if } \tilde{M}_{t-\tau} \leq 0.5 \end{cases}$$

Secondly, learned mask is binarised through threshold $T$ grid search which minimise:

$$|y_t^{(i)} - F^{(i)}(X_{t-\tau} \odot (\tilde{M}_{t-\tau} > T), \theta^{(i)})| + \lambda_3 \sum_{t=1}^{\tau} \sum_{j=1}^{N} |M_t^{(j)} > T_t^{(i)}| \to min$$

Learned binary mask represents importance of each timestamp of each time series from input multivariate set for forecasting next target value. From this mask dependency parameters are estimated:

$$\hat{\alpha}_{ij} = \begin{cases} 1, & \text{if } M_{t-\tau}^{(j)} \neq 0 \\ 0, & \text{if } M_{t-\tau}^{(j)} = 0 \end{cases}$$

$$\hat{\tau}_{ij} = \{t \mid M_t^{(j)} \neq 0\}$$

# 5. Experiments

We conducted following experiments on synthetic and real-world datasets:

## 5.1 Synthetic linear dependencies

Multivariate time series was generated with following linear dependencies:

$$X_t^{(i)} = \sum_{j \in R_i} \alpha_{ij} X_{t-\tau_{ij}}^{(j)} + \epsilon \sim N(0,1), i = 1,...,N$$

where
$$N \sim Uniform[5,15]$$
is the number of time series in set,
$$R_i = \{k_l \mid k_l \sim Uniform[0,N]\}, l = 1,...,n_i, n_i \sim Uniform[0,N]$$
is set of regressors which defines dynamics of $X_t^{(i)}$,
$$\alpha_{ij} \sim Uniform[-0.4, 0.4], j \in R_i$$
is set of coefficients which regulates strength of dependency,
$$\tau_{ij} \sim Uniform[0, 250], j \in R_i$$
is set of time lags which defines temporal dependency in time series. We generate 10 time series sets according following rules, randomly choose target time series and reveal dependencies in it us-ing described above importance estimation procedure and deep neural networks architectures. Quality of importance estimation is measured in terms of precision and recall. Example of generated set of time series:

## 5.2 Synthetic nonlinear dependencies

Following system of equations with nonlinearity is considered:

$$X_t^{(0)} = \sin(0.5t)\cos(2t) + \epsilon \sim N(0,1)$$
$$X_t^{(1)} = \sin(2t) + \cos(0.5t) + \epsilon \sim N(0,1)$$
$$X_t^{(2)} = X_{t-10}^{(0)} X_{t-100}^{(1)} + X_{t-70}^{(0)} X_{t-40}^{(1)} + \epsilon \sim N(0,1)$$
$$X_t^{(3)} = X_{t-150}^{(2)} X_{t-20}^{(1)} - 5\sin(X_{t-100}^{(0)}) + 0.1\epsilon \sim N(0,1)$$
$$X_t^{(4)} = X_{t-80}^{(3)}/(20 + X_{t-40}^{(2)}) + 0.1\epsilon \sim N(0,1)$$
$$X_t^{(5)} = X_{t-10}^{(0)} X_{t-20}^{(1)} + X_{t-110}^{(2)} X_{t-120}^{(4)} + X_{t-210}^{(4)} X_{t-270}^{(5)} + 0.1\epsilon \sim N(0,1)$$

We examine dependencies for $X_t^{(2)}, X_t^{(3)}, X_t^{(4)}, X_t^{(5)}$.

## 5.3 Real world dataset

We chose SML2010 dataset for estimating temporal dependencies. SML2010 is public dataset where target is forecast indoor temperature based on external sensors. Target variable was 'Temperature_Comedor_Sensor' and we built regression model based on 19 external sensors.

## 5.4 Settings

For synthetic datasets we generate 100000 data points, using first 80000 as train and last 20000 as test. We use Adam optimiser with learning rate 0.001 for all regression tasks. We fixed number of importance estimation steps to 20, importance learning rate to 0.1, importance regularisation coefficients $\lambda_1 = 0.0005, \lambda_2 = 0.5, \lambda_3 = 0.00001$ for nonlinear synthetic set and
$\lambda_1 = 0.005, \lambda_2 = 0.5, \lambda_3 = 0.0001$ for linear cases, following [6]. Precision and recall are calculated based on estimation of timestamps and variables importance map. For details, see https://github.com/KurochkinAlexey/discovering-long-term-dependencies-in-noisy-time-series-data.

# 6. Results and discussion

## 6.1 Synthetic linear dependencies

Results of revealing synthetic linear temporal dependencies in terms of precision and recall are summarised separately for RNN based architectures and TCN variants in Table 1 and Table 2:

Table 1. Results for RNN based architectures

|  | LSTM | GRU | Antisymmetric RNN | IMV LSTM | RHN |
|---|---|---|---|---|---|

|        | Precision | Recall | Precision | Recall | Precision | Recall | Precision | Recall | Precision | Recall |
|--------|-----------|--------|-----------|--------|-----------|--------|-----------|--------|-----------|--------|
| Case 1 | 0 | 0 | 1 | 0.125 | 0 | 0 | 1 | 0.125 | 0 | 0 |
| Case 2 | 0 | 0 | 0 | 0 | 0 | 0 | 0 | 0 | 0 | 0 |
| Case 3 | 0 | 0 | 0 | 0 | 0 | 0 | 0 | 0 | 0 | 0 |
| Case 4 | 0 | 0 | 0 | 0 | 0 | 0 | 0 | 0 | 0 | 0 |
| Case 5 | 0 | 0 | 0 | 0 | 0 | 0 | 0 | 0 | 0 | 0 |
| Case 6 | 0 | 0 | 0 | 0 | 0 | 0 | 0 | 0 | 0 | 0 |
| Case 7 | 0 | 0 | 0 | 0 | 0 | 0 | 0 | 0 | 0 | 0 |
| Case 8 | 0 | 0 | 0 | 0 | 0 | 0 | 0 | 0 | 0 | 0 |
| Case 9 | 0 | 0 | **0.5** | 0.091 | 0 | 0 | 0.143 | 0.091 | 0 | 0 |
| Case 10 | 0 | 0 | 0 | 0 | 0 | 0 | 0 | 0 | 0 | 0 |

Table 2. Results for TCN based architectures

|        | TCN | | Output attention TCN | | Layerwise attention TCN | | TCN Stack | | Bidirectional TCN | |
|--------|-----------|--------|-----------|--------|-----------|--------|-----------|--------|-----------|--------|
|        | Precision | Recall | Precision | Recall | Precision | Recall | Precision | Recall | Precision | Recall |
| Case 1 | 1 | 0.625 | 0 | 0 | 0.8 | 0.5 | 1 | **0.875** | 1 | 0.625 |
| Case 2 | 1 | 0.333 | 0 | 0 | 1 | 0.333 | 1 | **0.667** | 0.5 | 0.333 |
| Case 3 | 1 | 0.429 | 0 | 0 | 1 | 0.357 | 0.888 | **0.571** | 1 | 0.5 |
| Case 4 | 1 | 0.333 | 0 | 0 | 1 | 0.333 | 0.667 | **0.667** | 1 | 0.333 |
| Case 5 | 0 | 0 | 0 | 0 | 0 | 0 | **1** | **1** | 0 | 0 |
| Case 6 | 1 | 0.4 | 0 | 0 | 1 | 0.2 | 1 | **0.6** | 0.75 | 0.6 |
| Case 7 | 0 | 0 | 0 | 0 | 0 | 0 | 1 | 1 | 1 | 1 |
| Case 8 | 1 | 0.375 | 0 | 0 | 0.5 | 0.25 | 1 | **0.875** | 1 | 0.5 |
| Case 9 | 1 | 0.182 | 0 | 0 | 1 | 0.182 | 1 | **0.545** | 1 | 0.273 |
| Case 10 | 1 | 0.25 | 0 | 0 | 1 | 0.5 | 1 | 0.5 | 1 | 0.25 |

As can be seen from Table 1, recurrent neural network based architectures and output attention TCN are not capable of capturing long term dependencies in noisy time series data. Thus, we continue revealing dependencies in nonlinear datasets only with TCN based architectures except output attention TCN. We can observe that TCN Stack architecture outperforms others in terms of resulted recall. Notice, that many dependencies are not revealed because of low linear coefficient in target equation. We can conclude that different architectures of TCN can outperform default one. Also it was found that ability of network to find dependencies is essentially connected with network ability to train and generalise on test data. See train/test loss values in https://github.com/KurochkinAlexey/discovering-long-term-dependencies-in-noisy-time-series-data.

## 6.2 Synthetic nonlinear dependencies

Results of revealing nonlinear temporal dependencies are summarised in Table 3.

Table 3. Results for TCN based architectures

|  | TCN | | Layerwise attention TCN | | TCN Stack | | Bidirectional TCN | |
| --- | --- | --- | --- | --- | --- | --- | --- | --- |
|  | Precision | Recall | Precision | Recall | Precision | Recall | Precision | Recall |
| $X^{(2)}$ | 1 | 1 | 1 | 1 | 0 | 0 | 1 | 1 |
| $X^{(3)}$ | 0 | 0 | 0 | 0 | 1 | 0.667 | 1 | **1** |
| $X^{(4)}$ | 1 | 0.5 | 1 | 0.5 | 1 | 0.5 | 1 | 0.5 |
| $X^{(5)}$ | 1 | 0.333 | 1 | 0.333 | 0.833 | **0.833** | 1 | 0.333 |

We can observe that TCN Stack and Bidirectional TCN are leaders in terms of precision and recall. Also it was found that non of considered networks were able to detect 'division' nonlinearity in $X^{(4)}$ equation

## 6.3 SML2010

As there are no ground truth labels for SML2010 dataset, results can not be assessed quantitatively. Figure 2 shows resulted importance maps for 3 runs with different network parameters initialisation of default TCN architecture

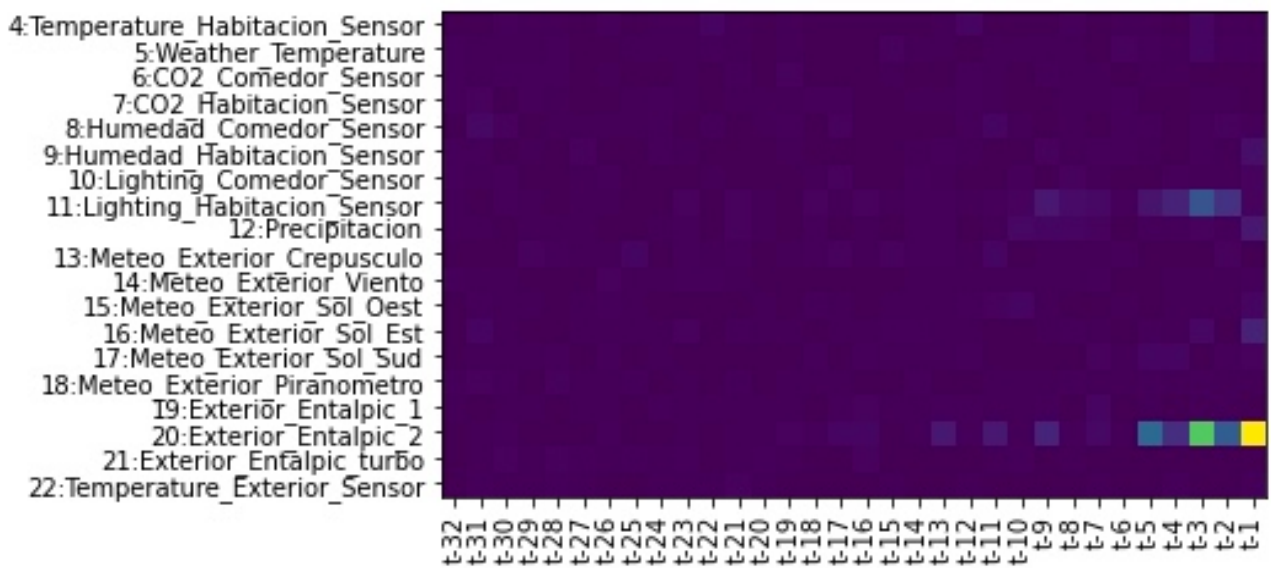

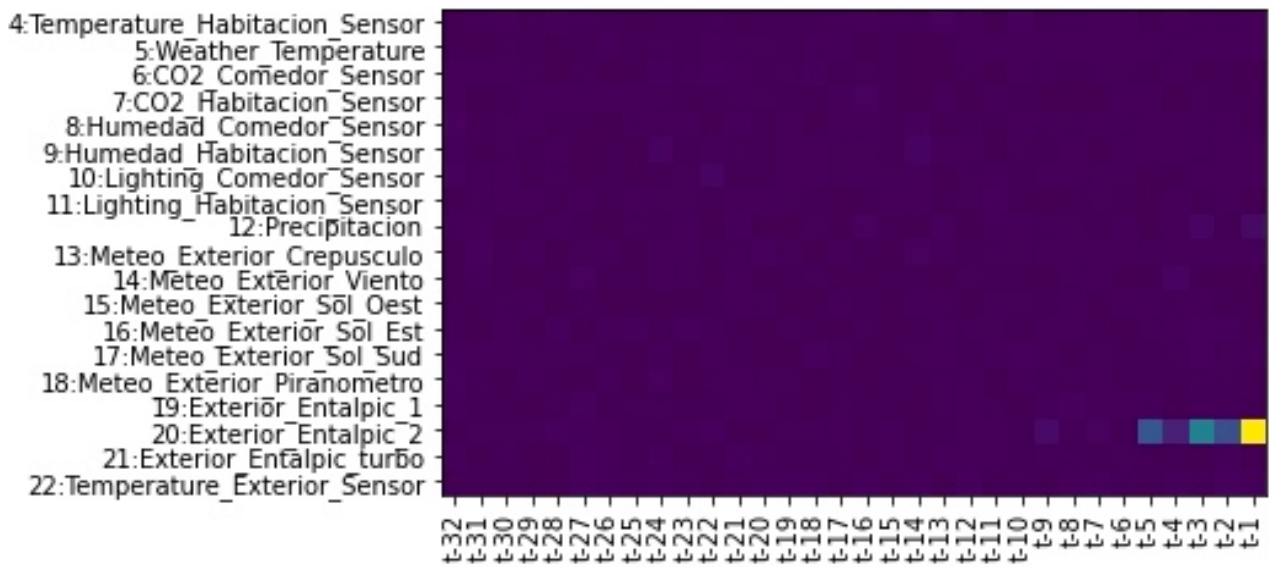

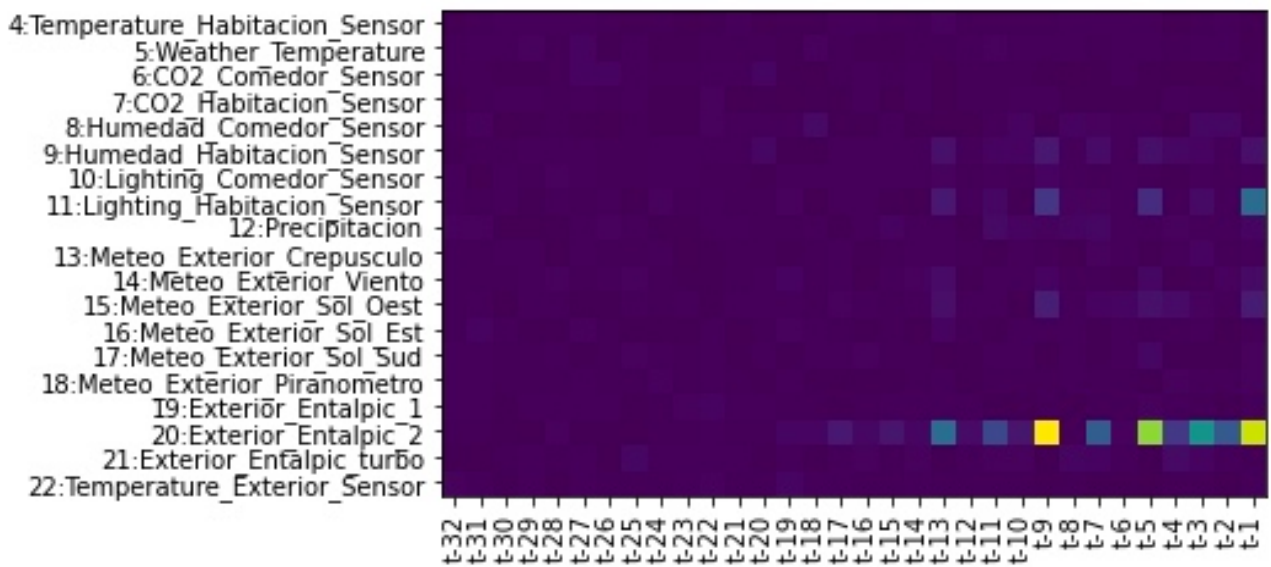

Figure 2. Resulted importance maps for 3 runs of TCN architecture on SML2010 dataset

It was found that different architectures produce different importance maps. We can conclude that on real world datasets:
1) Temporal and variable dependencies are not static like in our experiments with synthetic datasets, but dynamic and thus average importance map on these datasets is useless. One should assess temporal dependencies for given timestamp, not on whole dataset.
2) Results of revealing temporal and variable dependencies depend on local minima of loss landscape for which neural network optimisation converge.

## 7. Conclusion

We developed framework which is capable of revealing temporal and variable linear and nonlinear dependencies in noisy multivariate time series data. We considered different state-of-the-art recurrent neural network based architectures and different temporal convolutional network architectures and found that only TCN variants are able to found dependencies is this sort of data. In real world datasets, dependencies are dynamic and should be estimated for given timestamp. Also, results of this estimation depend on local minima for which neural network optimisation converge. It is an open question how to choose regularisation parameters for learned binary masks estimation, because it strongly affects results of importance calculation. Different sets of regularisation parameters can lead to very different results for low and high dimensional cases. It was also found that ability to capture long term dependencies is strongly connected with neural network ability to generalise on test set.